\documentclass{article}

\usepackage{PRIMEarxiv}
\usepackage{multirow}
\usepackage{makecell}
\usepackage{amsmath}
\usepackage{afterpage}
\usepackage{rotating}

\usepackage[utf8]{inputenc} 
\usepackage[T1]{fontenc}    
\usepackage{hyperref}       
\usepackage{url}            
\usepackage{booktabs}       
\usepackage{amsfonts}       
\usepackage{nicefrac}       
\usepackage{microtype}      
\usepackage{lipsum}
\usepackage{graphicx}
\graphicspath{{media/}}     
\usepackage{float}
  
\title{GAMedX: Generative AI-based Medical Entity Data Extractor Using Large Language Models

}

\author{
  Mohammed-Khalil Ghali, Abdelrahman Farrag, Hajar Sakai, Hicham El Baz, Yu Jin, Sarah Lam \\
  School of Systems Science and Industrial Engineering \\
  State University of New York at Binghamton \\
  Binghamton, NY, USA\\
  \texttt{mghali1, afarrag1, hsakai1, helbaz1, yjin, sarahlam@binghamton.edu} \\
}

\begin{document}
\maketitle

\begin{abstract}
In the rapidly evolving field of healthcare and beyond, the integration of generative AI in Electronic Health Records (EHRs) represents a pivotal advancement, addressing a critical gap in current information extraction techniques. This paper introduces GAMedX, a Named Entity Recognition (NER) approach utilizing Large Language Models (LLMs) to efficiently extract entities from medical narratives and unstructured text generated throughout various phases of the patient hospital visit. By addressing the significant challenge of processing unstructured medical text, GAMedX leverages the capabilities of generative AI and LLMs for improved data extraction. Employing a unified approach, the methodology integrates open-source LLMs for NER, utilizing chained prompts and Pydantic schemas for structured output to navigate the complexities of specialized medical jargon. The findings reveal significant ROUGE F1 score on one of the evaluation datasets with an accuracy of 98\%. This innovation enhances entity extraction, offering a scalable, cost-effective solution for automated forms filling from unstructured data. As a result, GAMedX streamlines the processing of unstructured narratives, and sets a new standard in NER applications, contributing significantly to theoretical and practical advancements beyond the medical technology sphere.
\end{abstract}


\section{Introduction}
The integration of Artificial Intelligence (AI) in healthcare, particularly through Electronic Health Records (EHRs), marks a significant advancement in medical technology. This progression is essential for enhancing healthcare delivery and improving patient outcomes, aiming at efficiently extracting and analyzing patient information from EHRs, which contain a blend of structured data such as coded diagnoses and medications and unstructured data, including clinical narratives and notes. While structured data entry in EHRs offers numerous benefits and is increasingly prevalent, its practical use by clinicians remains limited due to the added documentation burden \cite{bush2017structured}. Consequently, healthcare providers often prefer documenting patient information through clinical narratives \cite{meystre2008extracting}. These narratives, rich in detailed patient information, are crucial for enhancing the accuracy of diagnostic and prognostic models \cite{liang2019evaluation, yang2021assessing}. However, the free-text format of these narratives poses a significant challenge: they are not readily amenable to computational analysis, which typically requires structured data. This challenge is further compounded by the intrinsic complexities of clinical text, including irregularities like ambiguous medical jargon and nonstandard phrase structures. Despite the powerful capabilities of Natural Language Processing (NLP) to comprehend medical language in healthcare settings \cite{nadkarni2011natural}, such irregularities make it difficult for standard NLP tools to perform effectively when applied to clinical text, which necessitates domain-specific expertise for accurate annotation \cite{zheng2011coreference}.
However, the integration of Large Language Models (LLMs) into the healthcare sector is not without its constraints, particularly due to the confidentiality requirements governing clinical information. These requirements significantly restrict the availability and utilization of public datasets, which are essential for training and fine-tuning LLMs. This constraint is further compounded by the need for secure and compliant IT system integration in healthcare \cite{drolet2017electronic}. The sensitivity of patient data requires robust security measures to prevent unauthorized access and ensure data privacy \cite{reddy2023evaluating}. Additionally, healthcare IT systems often involve complex, diverse software ecosystems, requiring LLMs to be adaptable and interoperable with various existing platforms and data formats \cite{yang2023large}. This results in a limited supply and restricted distribution of these resources, leading to the creation of clinical NLP datasets that are limited and institution-specific \cite{xia2012clinical}. Each healthcare institution tends to possess unique, domain-specific data that is distinct from data held by other institutions. Consequently, this situation gives rise to a collection of diverse and institution-specific datasets, complicating the development of broadly applicable NLP tools in the healthcare field. These models that do not integrate these elements are generally restrained to tasks where labels are naturally generated in the course of clinical practice, such as the prediction of International Classification of Diseases Codes \cite{zhang2020bert} or mortality risk assessments \cite{si2019deep}.

\begin{figure}[H]
    \centering
    \includegraphics[width=1.0\textwidth]{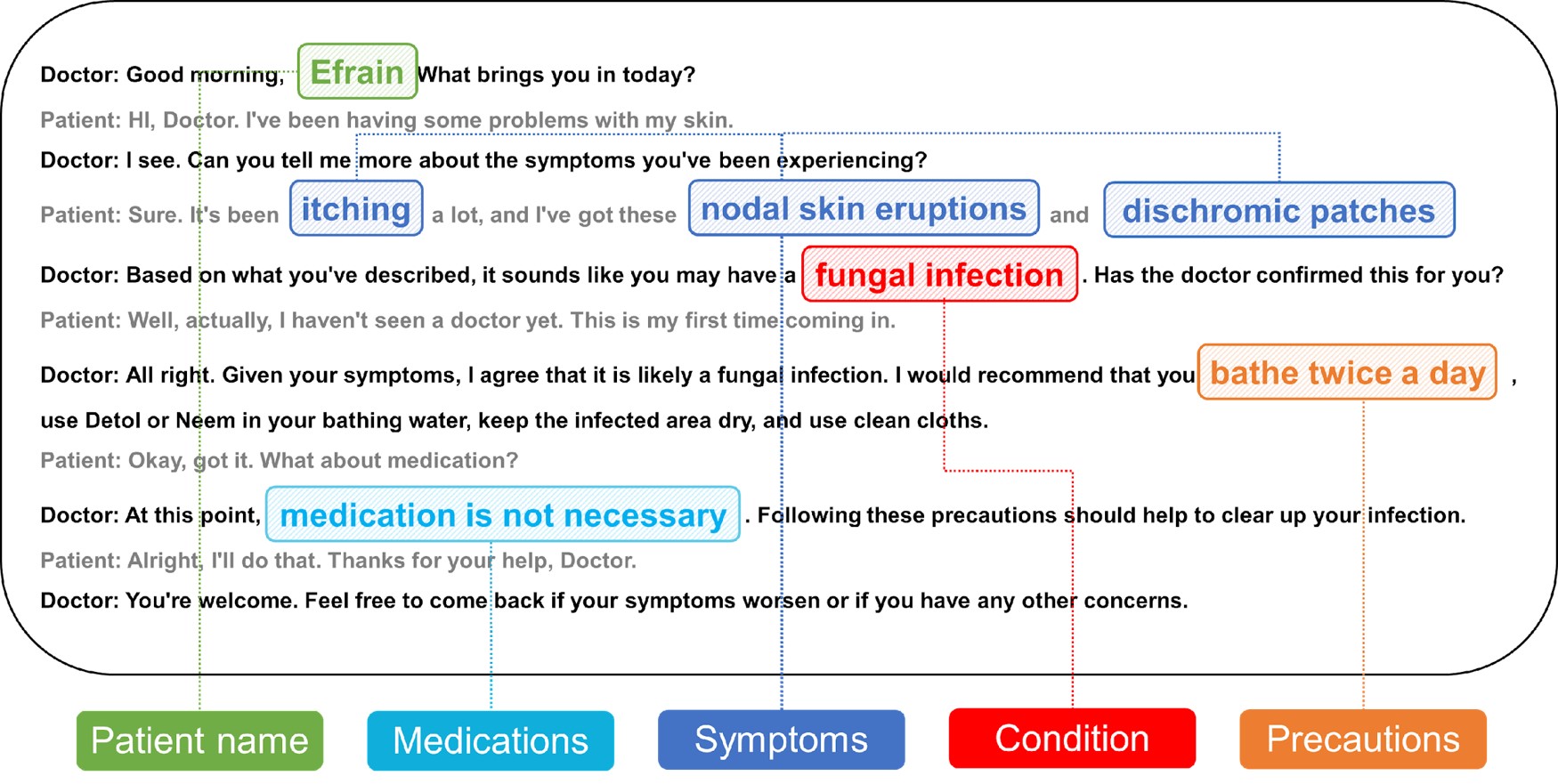}
    \caption{Example of a patient-doctor dialogue with annotated data elements for NER, highlighting the extraction of patient names, medications, symptoms, conditions, and precautions.}
    \label{fig 1}
\end{figure}

In response to these challenges, there is an emerging trend and a pressing need to develop AI-enabled, next-generation Generative Pretrained Transformer (GPT) or LLMs specifically tailored for the healthcare industry \cite{chakraborty2024need}. These advanced models not only provide accurate and error-free medical information but also address the ethical, legal, and practical concerns inherent in their deployment in sensitive healthcare environments. This paper proposes developing GAMedX, an advanced Named Entity Recognition (NER) system utilizing LLMs. The system is designed to precisely extract essential information from medical conversations and dictations. The intended outcome is a marked improvement in the efficiency of completing structured healthcare forms, with an emphasis on reliability, consistency, and a seamless operational workflow. GAMedX is extended beyond the technological integration, including critical real-world impacts such as accuracy, processing speed, user satisfaction, compliance, and the smooth integration of the solution into existing healthcare systems. This paper is structured as follows: Section 2 reviews related literature on the integration of LLMs in the healthcare sector for information extraction, with a focus on prompt engineering and pretrained LLMs for clinical NLP applications. Section 3 details the proposed novel model. Section 4 presents the evaluation results of this model. Finally, the conclusion and directions for future work are discussed in Section 5.

\section{Literature review}

The integration of AI in the healthcare domain has been significantly advanced by developments in NLP. Most NLP solutions leverage deep learning models \cite{lecun2015deep} based on neural network (NN) architectures, a rapidly evolving area within machine learning. Initially, Convolutional Neural Networks (CNNs) \cite{collobert2011natural} and Recurrent Neural Networks (RNNs) \cite{lample2016neural} were employed in early deep learning applications for NLP. They struggled with processing long-term dependencies and contextual information in large text sequences \cite{vaswani2017attention}. However, transformer architectures, notably the Bidirectional Encoder Representations from Transformers (BERT) \cite{devlin2018bert}, have recently set a new standard in NLP. These models are distinguished by their self-attention mechanism \cite{vaswani2017attention}, which efficiently processes the relative significance of each word in a sentence, enhancing understanding of context and relationships within the text. This capability has led to transformers overperforming other models in various NLP tasks. For instance, in NER \cite{yu2020named, yamada2020luke}, key entities in the text were identified and categorized, such as names of people, organizations, or locations; relation extraction transformers \cite{lyu2021relation, ye2021packed, xu2021entity} discern and extract relationships between entities within a text; sentence similarity tasks \cite{raffel2020exploring, jiang2019smart, yang2019xlnet} involve evaluating the degree of similarity or relatedness between two sentences; natural language inference \cite{zhang2020semantics, lan2019albert} is about determining the logical relationship between a pair of sentences, such as whether one implies, contradicts, or is neutral to the other; question answering \cite{zhang2021retrospective, garg2020tanda} these models comprehend a given text and accurately respond to questions based on that text, demonstrating a deep understanding of content and context.
The healthcare and medical sectors are facing the challenge of streamlining medical documentation processes, which are essential but also labour-intensive and time-consuming. Addressing this issue has increased interest in LLMs to develop improved NER systems. These systems are designed not only to extract and interpret information from medical dialogues and dictations accurately but also to efficiently summarize this information. However, general transformer models such as GPT often struggle to extract accurate information due to their training on more general datasets rather than specialized healthcare data. Additionally, LLMs have been shown to have significant drawbacks, such as generating misinformation, falsifying data, and contributing to plagiarism \cite{thirunavukarasu2023large}. The challenges extend further when implementing NER across different languages, each with its unique linguistic features and complexities. For example, Chinese medical text NER faces unique challenges, such as intricate terminology, variable entity lengths, and context-dependent entity classifications \cite{zhang2020using}. These concerns are particularly severe in the healthcare context, where the accuracy of information is paramount. Therefore, based on the literature, the development of such systems may follow one of three approaches: the first involves building LLMs that are specifically trained on healthcare data; the other two approaches involve adapting pre-existing models to clinical text through prompt engineering techniques or tuning specific layers, which can help guide the models to better understand and process medical-specific language and terms. 

\begin{figure}[H]
    \centering
    \includegraphics[width=1.0\textwidth]{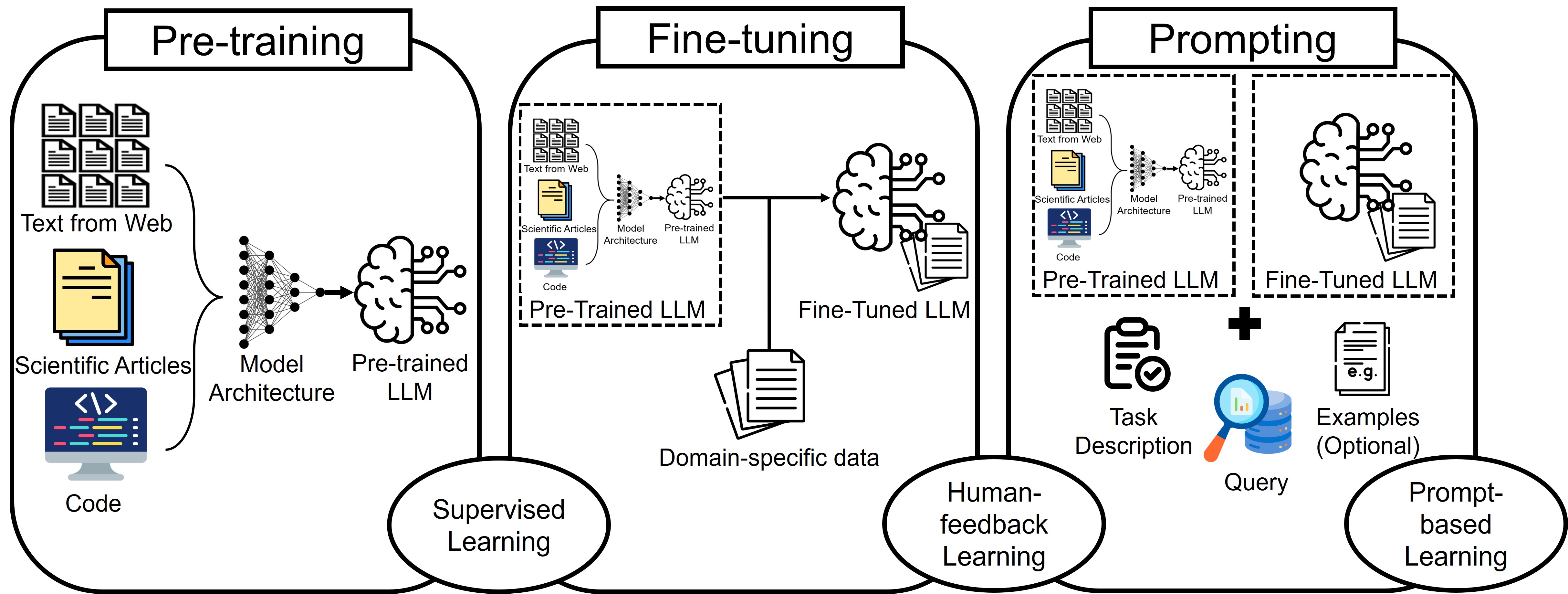}
    \caption{Overview of LLM development methods – Pre-Training on diverse sources, Fine-Tuning, and Prompting.}
    \label{fig 2}
\end{figure}

Training transformer models specifically for the healthcare sector is essential due to the significant differences in syntax and vocabulary between clinical text and the text typically used in general NLP. Clinical text is replete with specialized terminology and unique sentence structures that general language models may not recognize or understand properly \cite{wu2020deep}. Furthermore, the vastness of human language, with its nearly limitless combinations of words, sentences, meanings, and syntax, necessitates models that can comprehend and generate language with a high degree of accuracy \cite{bommasani2021opportunities}. The process of training these transformers typically occurs in two stages. The first is language model pretraining, where the model learns from a large corpus of unlabeled text, gaining an understanding of language through self-supervised learning objectives. The second stage is fine-tuning, where the model is refined to perform specific tasks using labeled training data. This process, known as transfer learning, allows the application of a model trained on one task to be adapted for another, leveraging the broad linguistic knowledge it has acquired. Recent studies have highlighted the superiority of large transformer models trained on massive text corpora over their predecessors, noting their enhanced language understanding and generation abilities. The significance of transformer models has directed research into extensive models, such as the GPT-3 \cite{floridi2020gpt}, which boasts 175 billion parameters and is trained on over 400 billion words of text, demonstrating remarkable performance. In the biomedical domain, specialized models like BioBERT \cite{Gu2021Biomedical} and PubMedBERT \cite{Gu2021Biomedical}, each with 110 million parameters, have been trained on PubMed's biomedical literature to better capture the language used in that field. NVIDIA has also developed the BioMegatron models, ranging from 345 million to 1.2 billion parameters \cite{shin-etal-2020-biomegatron}, using an extensive collection of text derived from PubMed. However, scaling transformer models in the clinical domain has been limited. This is partly due to the sensitive nature of clinical narratives, which often contain Protected Health Information, and the substantial computational resources required to train larger models. For instance, ClinicalBERT \cite{alsentzer-etal-2019-publicly}, a substantial model with 110 million parameters, was developed using the MIMIC-III dataset, which includes 0.5 billion words of clinical narratives and stands as one of the largest models tailored for the clinical domain. Yet, recent advancements have led to the emergence of GatorTron \cite{yang2022large}, which represents a significant evolution in clinical language models. Trained with 8.9 billion parameters on a corpus exceeding 90 billion words, including 82 billion from de-identified clinical texts, GatorTron showed significant improvements in model scale and performance on clinical NLP tasks. However, this approach is burdened by substantial challenges. The development and training of LLMs demand extensive computational resources, necessitating significant financial investments \cite{arora2023promise}.
\\[0.25cm]
Additionally, for effective training, LLMs require vast amounts of data, often meaning that healthcare institutions must collaborate and share data. This aspect poses a particular challenge in the healthcare sector, where strict data privacy regulations and institutional data protection agreements are in place. Such constraints can lead to the creation of data silos, impeding the free exchange of information necessary for training these models \cite{arora2023promise}. Furthermore, not every hospital or healthcare institution has the financial capacity to either rent these pretrained models or invest in training their own model.
\\[0.25cm]
Prompt-based learning is a technique where a pre-trained model is adapted to perform specific tasks by using carefully constructed text prompts. These prompts are designed to provide context and instruction, guiding the model in its response. Prompts are crucial in leveraging the model's extensive pre-existing language understanding for new, specific applications. The process of prompt-based learning typically involves two primary steps: designing the prompt involves creating a clear, concise textual instruction that explains the task to the model. The prompt sets the context and often includes specific keywords to guide the model's response in the desired direction, as well as in-context examples, in which prompts are supplemented with in-context examples. These examples serve as demonstrations of the task, showing the model the type of input, it will receive and the expected format of its response. The inclusion of examples varies; it could range from none “zero-shot”, relying only on the prompt's guidance, to several “few-shot”, providing a more comprehensive context. These approaches allow the LLMs to be quickly adapted to a wide range of general-domain tasks with little to no specific prior training \cite{mann2020language, liu2023pre, sanh2021multitask}. 
\\[0.25cm]
Prompt-based learning has shown significant progress in classification tasks such as multiple-choice questions \cite{mishra2021reframing}, demonstrating its adaptability even in complex scenarios like coreference resolution. In these instances, models are prompted to simplify the task by choosing between two potential antecedents for a pronoun or confirming the correctness of a single antecedent \cite{yang2022gpt}. This method's effectiveness relies heavily on the generation of a comprehensive list of potential antecedents, necessitating additional tools or multiple queries, potentially increasing computational demands. However, the literature reveals challenges in effectively dealing with information extraction tasks. These challenges include the extraction of multiple interconnected concepts essential for understanding patient information \cite{liu2022qaner}. Furthermore, there are issues with overlapped and nested concepts, where a single concept might be categorized under multiple labels or linked to various relations, complicating annotations \cite{yang2020identifying}. Efficiently extracting relations is challenging, as enumerating all concept combinations before classification leads to a skewed positive-negative ratio, given that only a few combinations have actual relations \cite{yang2019madex}. Additionally, the adaptability of concept and relation extraction methodologies across different institutional settings raises concerns about the portability and generalizability of the model, underscoring the need for strategies that can accommodate the diversity of clinical documentation practices \cite{yang2019study, ferraro2017effects}. To tackle the issue of effectively extracting clinical concepts and their relations, a unified prompt-based Machine Reading Comprehension (MRC) architecture is proposed, utilizing state-of-the-art transformer models \cite{peng2023clinical}. The study benchmarks its MRC models against existing deep learning models on two key datasets from the National NLP Clinical Challenges (n2c2) \cite{chen2023contextualized}: one focusing on medications and adverse drug events from 2018 \cite{henry20202018} and the other on the relations of social determinants of health (SDoH) from 2022 \cite{lybarger20232022}. However, the practice of human-designed, discrete prompts necessitates prior domain knowledge, limiting the method's generalizability across different NLP tasks without undergoing a similar process of "prompt-based learning" for each new task. The study's reliance on datasets derived from the publicly available Medical Information Mart for Intensive Care III (MIMIC-III) database raises concerns about data leakage, especially when utilizing models pre-trained on this same dataset. In addition, the approach is dependent on privately pre-trained models, which are not accessible for external validation or replication, further compounding these limitations. 
\\[0.25cm]
An approach was proposed to address this challenge by accessing the underlying model and using labeled data for training the extraction layer \cite{li2019unified}. However, this requirement can pose significant limitations in scenarios where such access is restricted or where labeled data is scarce or expensive to obtain. Another challenge is related to the output generated by the LLMs, such as those observed with InstructGPT \cite{ouyang2022training}, which, despite incorporating extraction examples into its training, fails to produce structured outputs. This limitation is significant as it necessitates additional steps to convert the model's text-based responses into a structured form that can be readily analyzed or integrated into existing healthcare systems. A handcrafted guided prompt design was proposed aiming at a unified output structure of LLMs for diverse clinical information extraction tasks, demonstrating versatility across zero- and few-shot learning scenarios \cite{agrawal2022large}. Despite its advancements, the model faces limitations, including difficulties in precisely matching detailed clinical schemas at the token level, a tendency to generate non-trivial answers even when none are required, and restrictions imposed by data use policies that limit the scope of training and evaluation datasets. Moreover, the model's primary reliance on English-language texts and data from dictated notes may not capture the full diversity of clinical documentation practices, posing challenges for its application across different linguistic and healthcare contexts.
\\[0.25cm]
To conclude the literature review, it is crucial to note that existing methodologies for extracting medical information predominantly involve training or fine-tuning LLMs to align with specific testing data requirements. A common strategy includes fine-tuning models; however, this often relies on proprietary, non-free, pre-trained medical LLMs and adopts a non-unified model approach \cite{ahmed2024med}. Such practices may limit accessibility, scalability, and the generalizability of solutions across different healthcare settings due to the proprietary nature of the models and the tailored approach to model training. In this study, GAMedX introduces an innovative wrapping approach designed to achieve a unified structure format for a NER system, focusing on the comprehensive understanding of multiple interconnected concepts to enhance patient information processing. GAMedX, leveraging open-source LLMs, offers a straightforward, cost-efficient solution. It is aimed at optimizing hospital resources and services, directing attention and funds more effectively towards improving patient health outcomes. GAMedX stands as a testament to the potential of harnessing advanced NLP technologies to not only advance clinical data processing but also to significantly contribute to the overall efficiency of healthcare delivery. 

\section{Data}

For the experiments conducted with the proposed approach, two datasets are considered. Each dataset contains textual data generated subsequent to medical encounters. Additionally, these datasets’ annotations are Named Entity Recognition (NER) task-oriented. The first dataset is a competition dataset that was synthetically generated by Prediction Guard. The competition is the Data 4 Good Challenge 2023 organized by Purdue University and during which we clinched the 1st place in the technical component consisting of using open-source LLMs for information extraction. The second dataset is extracted from Vaccine Adverse Event Reporting System (VAERS) database where the patient’s personal information is protected by being de-identified. As privacy is a key concern in healthcare and keeping the patient’s Protected Health Information (PHI) safe, it is important to mention that both datasets used for experiments are HIPAA compliant either because they were synthetically generated or a de-identification process was involved.  

\subsection{Dataset 1: Medical Transcripts (Data 4 Good Challenge)}

During a medical appointment, a conversation takes place between a patient and their provider. The medical concerns discussed are either recorded during the appointment or summarized and dictated by the provider afterward. As a result, multiple audio files are compiled. These files contain different patient’s medical information and are afterward transcribed into text data. The transcripts form a gold mine, however, being raw data, they end up being both challenging and time-consuming to retrieve information from. The information to be retrieved was highlighted in the example described in Figure \ref{fig 1}. 
\\[0.25cm]
This dataset was provided by Prediction Guard during the Data 4 Good challenge organized in Fall 2023. There are 2001 transcripts in this dataset and each of them corresponds to six labels. 

\subsection{Dataset 2: Vaccine Adverse Event Reporting System (VAERS)}

Jointly managed by both the US Centers for Disease Control and Prevention (CDC) and the U.S. Food and Drug Administration (FDA), the Vaccine Adverse Event Reporting System (VAERS) is set up for post-vaccination administration adverse events collection and analysis \cite{cdc_vaers}. As a result, uncommon patterns of adverse events can be identified and therefore indicate any potential safety issues with a specific vaccine. Therefore, this database provides an efficient tool to ensure a continued safety while administering vaccine by enabling early warning signals identifications and contributing to the public health decisions regarding vaccination recommendations. Additionally, this data is regularly updated and consists of a general comprehensive list of definitions, descriptions, and abbreviations. Similar to \cite{li2024ae}, 91 annotated safety reports are considered where the narratives are symptoms descriptions. The adverse events to extract are those related to a nervous system disorder in instances of Guillain-Barre Syndrome (GBS) associated with influenza vaccinations, as reported in the VAERS data. 

\begin{table}
    \centering
    \caption{Tasks and Datasets Description}
    \label{tab:results_summary}
    \begin{tabular}{ccccc}
    \toprule
\textbf{Task} & \textbf{Dataset} & \textbf{Description} & \textbf{Example} & \textbf{Extraction}  \\ \midrule
\multirow{2}{*}{\makecell{\textbf{Medical} \\ \textbf{Textual Data} \\ \textbf{Information} \\ \textbf{Extraction}}} & \makecell{Medical \\ Transcripts \\ (Data 4 Good \\ Competition) }& \makecell{Given a medical \\ transcript, \\ extract six \\ patient-related \\ information} & \makecell{"During my visit \\ with Ilana Bellinger, \\ an 85-year-old \\ patient who \\ presented […]"} & \makecell{- Name: Ilana Bellinger \\
- Age: 85} \\  \cmidrule(lr){2-5}
& \makecell{Vaccine \\ Adverse Event \\ Reporting \\ System \\ (VAERS)} & \makecell{Given a medical \\ report, extract \\ the post- \\ vaccination \\ adverse events} & \makecell{"Pt started feeling \\ dizzy immediately \\ after vaccine was \\ given […]"} & \makecell{Dizziness}  \\ \midrule

\end{tabular}

\end{table}

\section{Methodology}

A unified approach leveraging open-source pre-trained LLMs is proposed in this paper to extract structured information from healthcare textual data. Given the challenging nature of information extraction from unstructured data, an advanced Named Entity Recognition (NER) process was developed using a LLM wrapper. The goal is to automate redundant and time-consuming medical documentation and form-filling while ensuring a reliable system capable of being seamlessly integrated into the healthcare facility infrastructure. All this is, however, subject to one additional key consideration consisting of only harnessing open-source LLMs to ensure that no additional cost would be required to for smoothly run the process. The elaborated unified framework is summarized in Figure \ref{fig3}, based on which experiments were conducted and evaluations were carried out. 

\subsection{Loading and Preprocessing Data}
The first step consists of obtaining the dataset subject to experiments. In each dataset Dk, the document di is either a medical transcript or report and the output Oi consists of the extracted information:

\begin{equation}
D_k = \{(d_1, O_1), (d_2, O_2), \ldots, (d_m, O_m)\},
\end{equation}

where k = \{1,2\} refers to the dataset considered, and m the dataset size. 
\\[0.25cm]
The textual data compromises transcripts of patient-doctor conversations and dictations as well as medical reports. All documents can be either in ‘.txt’ or ‘.json’ format. Once loaded, they are transformed by using LangChain’s “Recursive text Splitter” in order to keep the semantically related chunks together while enabling batch processing imposed by the token limit and resorted to for computational efficiency. Moreover, if the document considered is not in English, it is translated into English:

\begin{equation}
T_i = P(d_i), 
\end{equation}
\noindent
where $T_i$ refers to the translated text and $P$ the preprocessing operation.

\begin{equation}
B_i = \sum_{j} T_i, 
\end{equation}
\noindent
where $B_i$ refers to the batch resulting from the “Recursive text Splitter” and $j = \{1, \ldots, J\}$ with $J$ being the number of batches.

\afterpage{%
    \clearpage
    \begin{sidewaysfigure}
        \centering
        \includegraphics[width=\textheight]{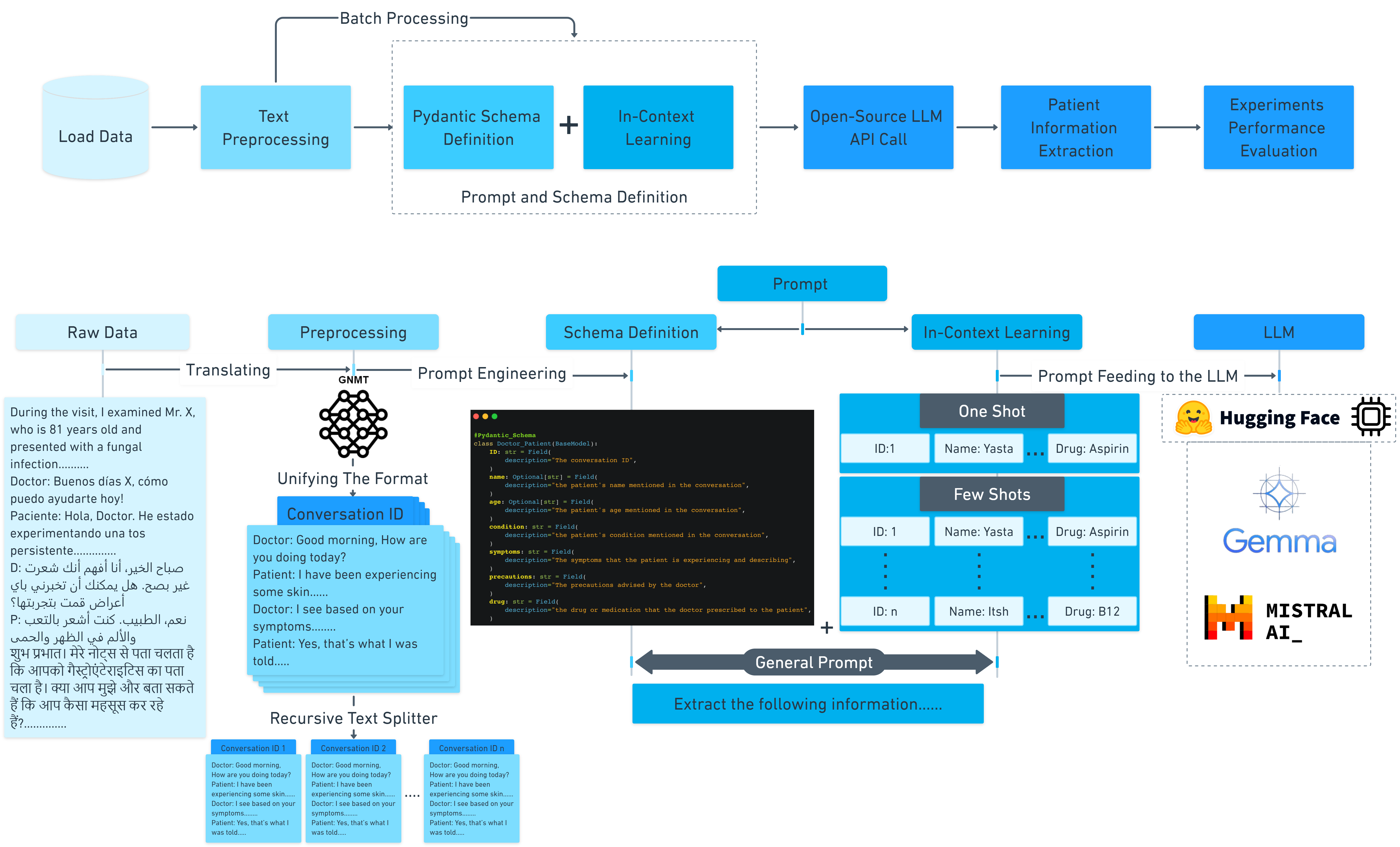} 
        \caption{Graphical abstract for the developed methodology: GAMedX}
        \label{fig3}
    \end{sidewaysfigure}
    \clearpage
}
\newpage

\subsection{Prompt Crafting \& Pydantic Schema}

In addition to the general prompt designed to extract the patient’s information or post-vaccination adverse events, a Pydantic Schema is established. It consists of defining the data type and format of the information to be retrieved. This ensures that the extracted data conforms to what is expected:

\begin{equation}
E(T_i) = ``\text{[Task Description]} \text{[Query: } T_i\text{]}", 
\end{equation}
\noindent
where $E(T_i)$ refers to the prompt engineering and pydantic schema process.

\begin{figure}[H]
\centering
        \includegraphics[totalheight=8cm]{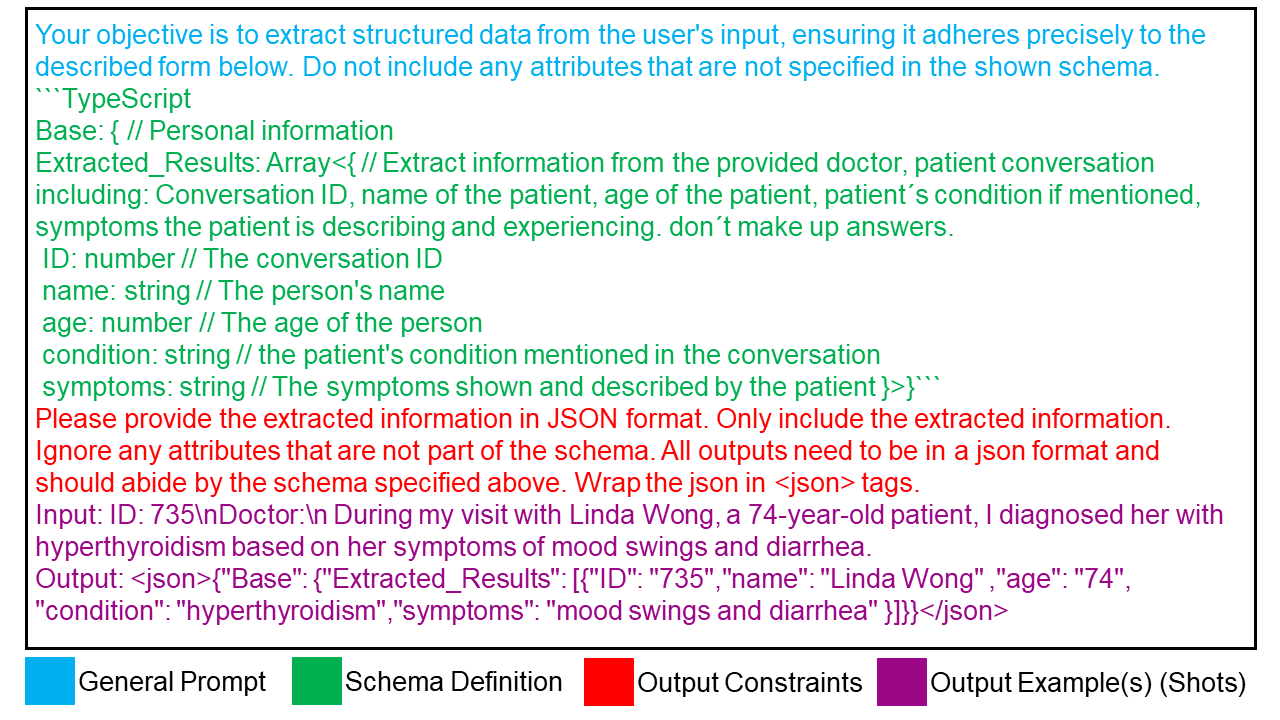}
    \caption{Example of the prompt used}
    \label{fig4}
\end{figure}

\subsection{Pre-trained Open-Source LLMs Used}

In this research the focus is to leverage open-source LLMs for healthcare-based information extraction. The selected models are Mistral 7B and Gemma 7B. Both these models are open models and therefore offer to developers the freedom to customize them and tailor them to specific downstream tasks. Additionally, being open source promotes both innovation and transparency and leverages AI democratization. For each of the LLMs considered, an API call is invoked with the complete prompt to extract the information intended: 

\begin{equation}
H_i = F(E(T_i)),
\end{equation}

where F represents the LLM API key call. 

\subsubsection{Mistral 7B}

Mistral 7B was released by Mistral AI under the Apache 2.0 license, permitting its use without restrictions \cite{jiang2023mistral}. It was designed to compete with larger language models in terms of performance with only 7.3 billion parameters. As a result, it surpasses Llama 2 13B on all benchmark, Llama 1 34B on multiple ones, while approaching Code Llama 7B on code-related tasks. Its architecture achieves faster inference thanks to Grouped-Query Attention (GQA) and can deal with longer sequences while minimizing the cost because of the Sliding Window Attention (SWA) integration \cite{jiang2023mistral}. Because this model is freely available, it is widely used by researchers and developers to build AI powered tools in a cost-effective manner. Moreover, Mistral 7B has demonstrated its good performance across various tasks such as Common Sense Reasoning, Arithmetic Reasoning, and Code Generation \cite{jiang2023mistral}, allowing it to be a potential candidate to be used to advance AI applications. 

\subsubsection{Gemma 7B}

Gemma 7B is part of Google’s Gemma family of LLMs. It is characterized with being lightweight and based on the same technology used for the Gemini models \cite{team2024gemma}. With only 7B parameters, this model introduces state-of-the-art AI capabilities. Its development is as the heart of Google’s responsible AI development strategy. Therefore, the pre-training was claimed to be conducted safely on well curated data, with a robust and transparent evaluation [65]. Gemma 7B demonstrated superior performance in multiple benchmarks such as Massive Multitask Language Understanding (MMLU) and HellaSwag and this shows its problem-solving ability [65]. Its capabilities are versatile, excelling in reasoning, math, and code tasks \cite{team2024gemma}.
\\[0.25cm]
Gemma’s technical report \cite{team2024gemma} compares three open LLMs (LLaMA 2, Mistral, and Gemma) using multiple benchmark datasets, based on that, Figure \ref{fig4} was deduced. 

\begin{figure}[H]
\centering
        \includegraphics[totalheight=8cm]{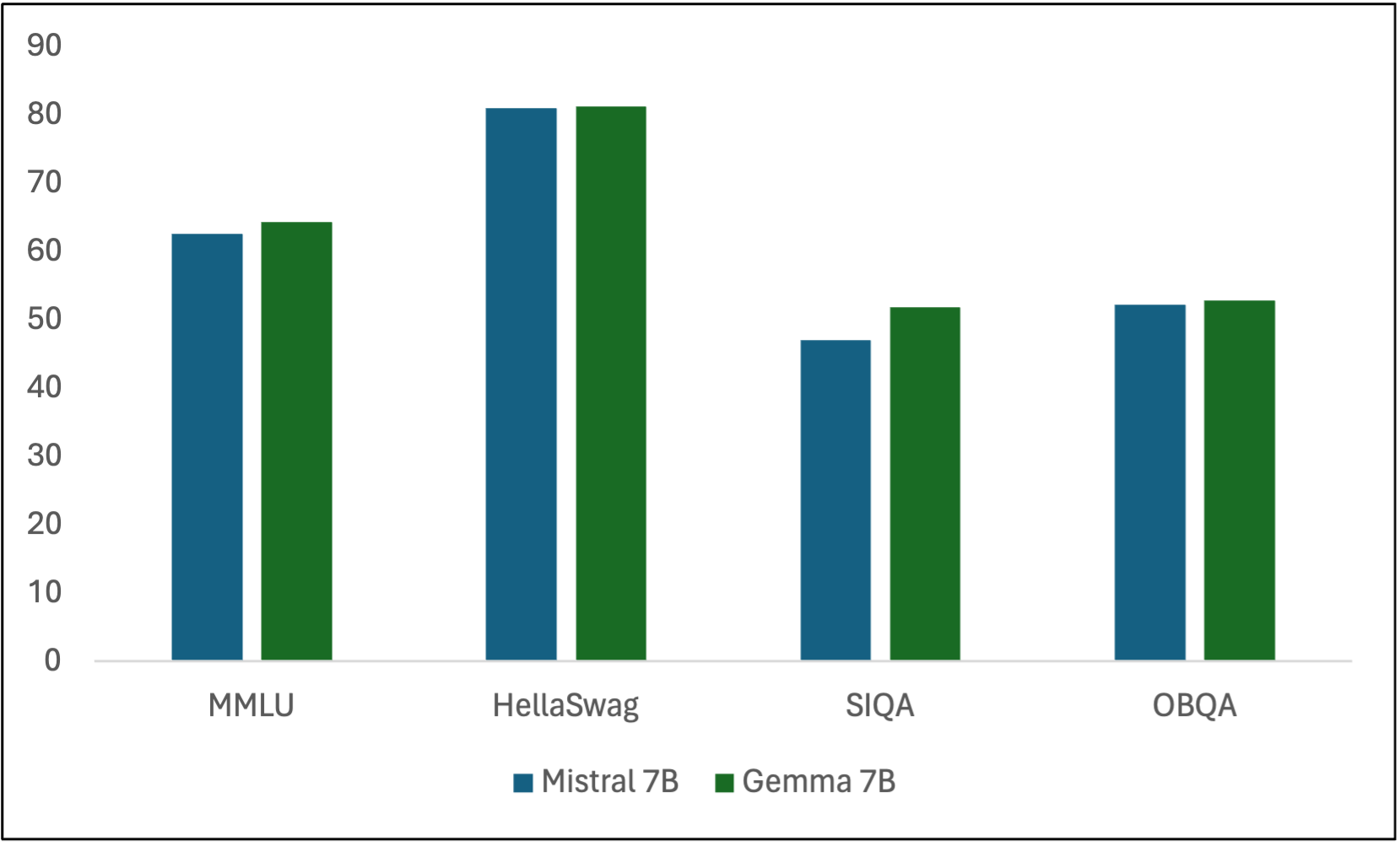}
    \caption{Benchmarks comparison of the LLMs used}
    \label{fig5}
\end{figure}

The benchmark datasets reported measure the ability of LLMs to conduct a variety of tasks. MMLU is used to evaluate the problem-solving skills, while HellaSwag is appropriate for common sense reasoning testing when it comes for generating the end of a story. Social Interaction QA (SIQA), on one hand, is reserved for evaluating the ability of the LLM to understand people’s social interactions and on the other hand OpenBookQA (OBQA) is suitable for measuring the LLM skill in carrying out advanced Question-Answering. This subset of comparison based on \cite{team2024gemma} shows that both Mistral 7B and Gemma 7B are very competitive across multiple datasets and tasks. 

\subsection{In-Context Learning}

In order to align the knowledge embedded in each of the pre-trained LLMs with the context provided in the prompt without explicitly re-training or fine-tuning on specific datasets for specific tasks, in-context learning was applied. This technique allows the adaptation of the LLM to the downstream tasks, in the case of this research, Medical Transcripts Information Extraction and Medical Reports Post-Vaccination Adverse Events Retrieval. As a result, a more efficient yet not computationally expensive LLM-application is enabled. Few-shots learning was conducted by providing examples in the prompt and allowing the model to have a concrete idea about the expected output. In the literature, few-shot learning commonly involves a range of examples varying between two and five. Based on this, multiple experiments were conducted resulting in choosing a three-shots learning for both Mistral 7B and Gemma 7B. 
\\[0.25cm]
The prompt is updated to the following in the case of in-context learning: 

\begin{equation}
E(T_i) = ``\text{[Task Description]} \text{[Examples]} \text{[Query: } T_i\text{]}",
\end{equation}

The experiments included from zero-shot to five-shots learning. However, only one-shot and few-shots learning are reported based on the results performance:

\begin{itemize}
    \item One-shot: Only one basic example is provided in the prompt.
    \item Few-shots: Two harder examples are added to the previous prompt.
\end{itemize}

\subsection{Performance Evaluation}
Post-processing is conducted to get the extracted information in the desired format. Following, two types of evaluations are carried out. The first evaluation is a Quantitative Analysis using ROUGE-1 F1 and ROUGE-L F1. ROUGE (Recall-Oriented Understudy for Gisting Evaluation) include a set of metrics commonly used to evaluate automatic text summarization by measuring the overlap on n-grams between the actual information extracted and the LLM output. In this paper, the focus is on two of its variants: ROUGE-1 and ROUGE-L where the first focuses on unigrams and the second on the longest common subsequence.  ROUGE-1 F1 and ROUGE-L F1 computations rely on the corresponding Precisions and Recalls.

\begin{table}[h]
    \centering
    \caption{Metrics and Formulas}
    \begin{tabular}{cc}
        \toprule
        \textbf{Metric} & \textbf{Formula} \\
        \midrule
        \textbf{ROUGE-1 Precision} & \(\frac{\text{Number of overlapping unigrams}}{\text{Total number of unigrams in the LLM output}}\) \\ \midrule
        \textbf{ROUGE-1 Recall} & \(\frac{\text{Number of overlapping unigrams}}{\text{Total number of unigrams in the actual information extracted}}\) \\ \midrule
        \textbf{ROUGE-1 F1} & \(\frac{2 \times (\text{ROUGE-1 Precision} \times \text{ROUGE-1 Recall})}{\text{ROUGE-1 Precision} + \text{ROUGE-1 Recall}}\) \\ \midrule
        \textbf{ROUGE-L Precision} & \(\frac{\text{Length of the longest common subsequence}}{\text{Total number of unigrams in the LLM output}}\) \\ \midrule
        \textbf{ROUGE-L Recall} & \(\frac{\text{Length of the longest common subsequence}}{\text{Total number of unigrams in the actual information extracted}}\) \\ \midrule
        \textbf{ROUGE-L F1 }& \(\frac{2 \times (\text{ROUGE-L Precision} \times \text{ROUGE-L Recall})}{\text{ROUGE-L Precision} + \text{ROUGE-L Recall}}\) \\ 
        \bottomrule
    \end{tabular}
\end{table}

However, for the VAERS dataset, conducting a Quantitative Analysis was revealed to not be enough. As a result, a Semantic Analysis is carried out. Two embedding models are used to plot the LLM outputs, and the corresponding actual information extracted using t-distributed Stochastic Neighbor Embedding (t-SNE). 
\\[0.25cm]
t-SNE is a dimensionality reduction technique \cite{van2008visualizing} particularly suited for high dimensional data visualization in a low dimensional space. The process is characterized by preserving the local structure of the data allowing efficient clusters reveal and complex patterns identification. Which is not the case of another popular dimension reduction technique, the Principal Component Analysis (PCA). Additionally, t-SNE is also capable of handling non-linear data contrary to PCA. 
\\[0.25cm]
The textual data is vectorized using two models. On one hand, the first model is BAAI General Embedding (BGE). It was introduced by Beijing Academy of Artificial Intelligence (BAAI) and is characterized by high-dimensional embeddings, versatility, and scalability. It is a cutting-edge approach that stood out by capturing the semantic nuances. On the other hand, the second model considered is Instruct Embedding. Since its training was done on diverse datasets and tasks, resulting in task-oriented representations where a contextual understanding is improved. The choice of two different models was mainly to avoid any bias that might be related to a particular model.
\\[0.25cm]
Another contrast of the LLM output and actual information extracted is done using the computation of Cosine Similarity. This metric is insightful since it allows the understanding of the semantic similarity in addition to visualize the relationships within high-dimensional data representing, in our case the LLM output and the actual information extracted.

\section{Results}
The prompt was carefully defined, and the in-context learning examples were meticulously selected for each of the datasets considered in this research. The experiments were conducted using both open-source LLMs, Mistral 7B and Gemma 7B. For all cases, the temperature was set to 0.1 in order to limit randomness and reduce the LLM creativity while the max number of tokens was fixed at 1000. Given the nature of the output, ROUGE-1 F1 and ROUGE-L F1 were picked to evaluate the performance of our proposed approach. Table 3 summarizes the resulting performance metrics.

\begin{table}[H] 
    \centering
    \caption{Results Summary}
    \label{tab:results_summary}
    \begin{tabular}{cccccc}
    \toprule
        \textbf{Model} & \textbf{Strategies} & \multicolumn{2}{c}{\textbf{Competition Dataset}} & \multicolumn{2}{c}{\textbf{VAERS Dataset}} \\ \midrule
        & & \textbf{ROUGE-1 F1} & \textbf{ROUGE-L F1}	& \textbf{ROUGE-1 F1} & \textbf{ROUGE-L F1} \\
        \midrule
        \textbf{Mistral} & \textbf{One Shot} & 97\% & 98\% & 58\% & 57\% \\
        \textbf{Mistral} & \textbf{Few Shots} & 98\% & 98\% & 63\% & 62\% \\
        \textbf{Gemma} & \textbf{One Shot} & 97\% & 97\% & 60\% & 59\% \\
        \textbf{Gemma} & \textbf{Few Shots} & 98\% & 98\% & 63\% & 62\% \\
        \bottomrule
    \end{tabular}
\end{table}

\subsection{Quantitative Analysis}

\subsubsection{Competition Dataset}

ROUGE-1 F1 and ROUGE-L F1 scores are significantly high across all models and experiments, nearly reaching the perfect score of 1. This indicates excellent performance in capturing the essence of the input content. Moreover, there is a consistent pattern of F1 scores across both the few-shot and one-shot experiments for each model, suggesting that the methodology is robust regardless of the amount of initial data provided or in-context learning examples included in the prompt. These results reinforce the conclusion that the methodology employed is highly effective for key information extraction in healthcare text analysis and is generalizable across different LLMs.

\subsubsection{VAERS Dataset}

In comparison to the competition dataset, the VAERS Dataset shows noticeably lower scores, ranging from 0.57 to 0.63 for both ROUGE-1 and ROUGE-L F1 scores. This suggests that the VAERS Dataset poses a tougher challenge for named entity recognition tasks, likely due to the complexity of medical terminology, diverse entity types, or less consistency in how entities are mentioned. Similar to what was observed in the Competition dataset, employing "Few Shot" learning leads to performance improvements for both models, although these improvements are not very significant. This indicates that providing a small number of targeted examples helps the models adapt to the specific characteristics of the VAERS Dataset, but it doesn't completely solve all the challenges.

\subsection{Semantic Analysis (Case of VAERS Dataset)}

Although ROUGE-1 and ROUGE-L F1 scores are commonly used to assess summarization performance, they have limitations. Particularly, they may not fully capture the semantic richness and contextual alignment of model outputs. This becomes apparent when examining the low ROUGE scores achieved on the VAERS dataset, where models are challenged to translate everyday language into technical medical terms. ROUGE scores, as seen in the previous analysis table, rely heavily on the exact match of word sequences between the reference summaries and model-generated ones. However, in the context of the VAERS dataset, where reference answers contain specialized medical terminology, these scores might not accurately reflect the model's ability to comprehend and express the nuanced meanings of symptoms described using different terms in the transcripts. To address this concern, we decided to conduct an additional analysis, for the case of VAERS dataset, using two different embeddings models: BGE and Instruct Embeddings. This approach allows us to capture the semantic relationships between the reference summaries and the model's responses more effectively.
\\[0.25cm]
The analysis is carried out by plotting the t-SNE plots for each of the textual dataset where the ground truth is contrasted with each of the LLMs strategies outputs all in their embedding’s formats. The t-SNE was used to decrease the dimensionality of the embeddings space to be able to visualize them in a 2D space. Furthermore, in an attempt to reduce the bias that might result from a single embeddings technique, theses assessments resorted to two embedding techniques: BGE and Instruct Embeddings. Additionally, given that these two embedding techniques highlight different underlying structures of the data, where BGE capture contextual information while Instruct form specific tasks or instructions-based clusters, nuanced insights can be deduced and therefore different angles to visualize similarities and differences within the data points can be analysed. 

\subsubsection{One-Shot Learning }
\begin{figure}[h]
\centering
        \includegraphics[totalheight=6cm]{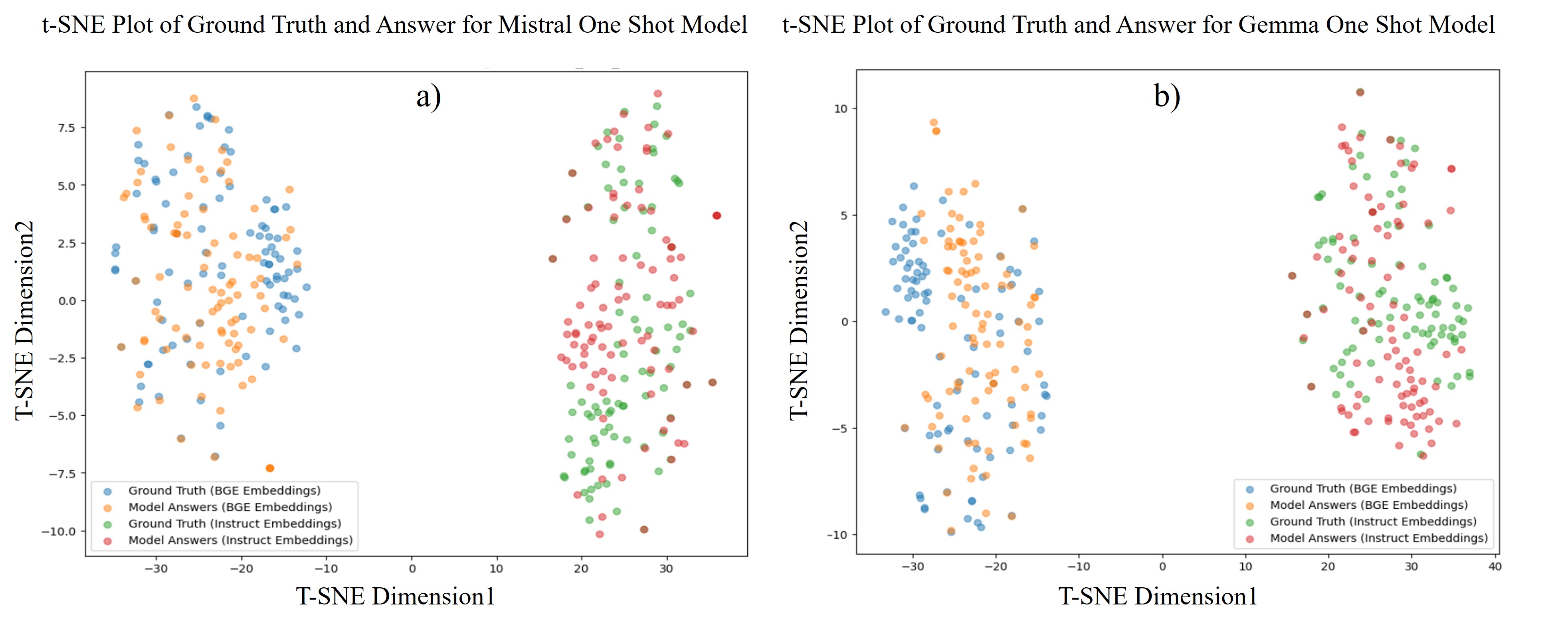}
    \caption{Figures 6.a and 6.b: t-SNE plot of ground truth and model answers for Mistral and Gemma using one shot prompt. }
    \label{fig6}
\end{figure}

Figure \ref{fig6} shows the t-SNE plots for both the Mistral and Gemma One-Shot models reveal a moderate level of clustering between the ground truth and our proposed methodology’s output using both embedding models' answers indicating a certain degree of semantic understanding captured by both models. Meanwhile, there is also a small dispersion in the clusters, suggesting instances where the models' interpretations diverge from the medical terminology employed in the ground truth. Nevertheless, the cosine similarity graphs support the assertion that both models provide relevant answers. Similarly, both models demonstrate overlap in clusters for ground truth and model answers with both types of embeddings. 
\begin{figure}[h]
\centering
        \includegraphics[totalheight=6cm]{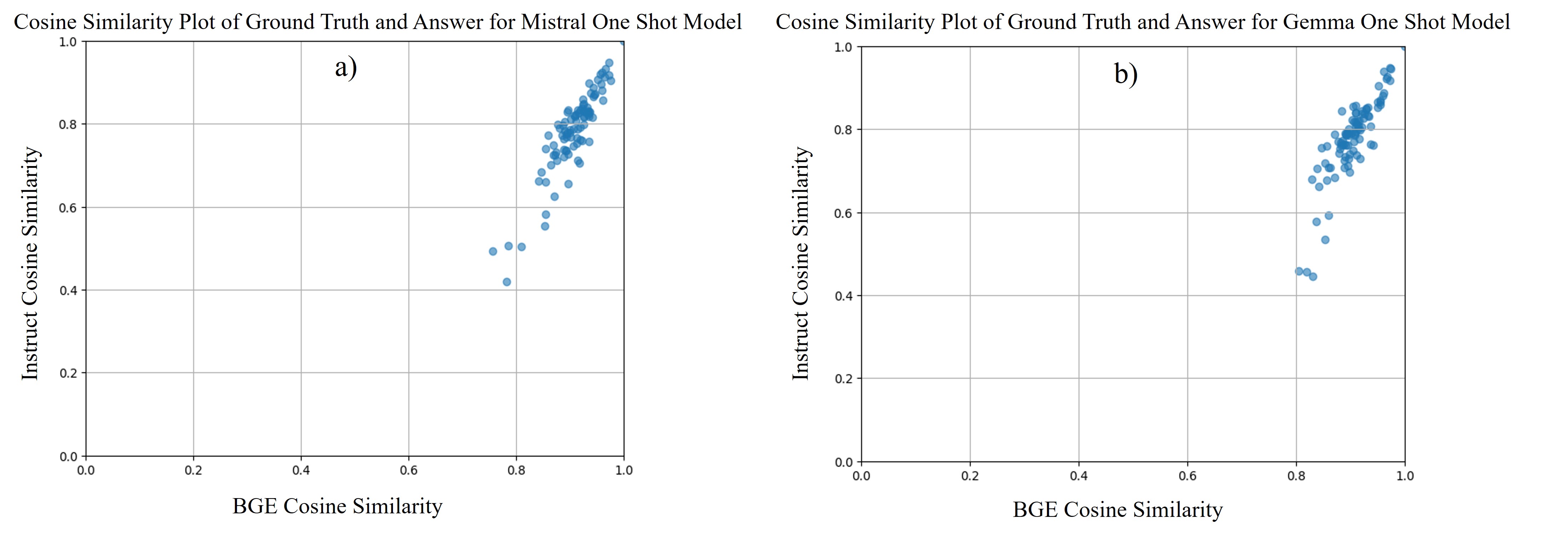}
    \caption{Figures 7.a and 7.b:  Cosine similarity plot for Gemma and Mistral models using one shot prompt.}
    \label{fig7}
\end{figure}

The cosine similarity graphs in figure \ref{fig7} also affirm the models' capability to grasp the semantic meaning of the symptoms.

\subsubsection{Few-Shot Learning}

\begin{figure}[h]
\centering
        \includegraphics[totalheight=6cm]{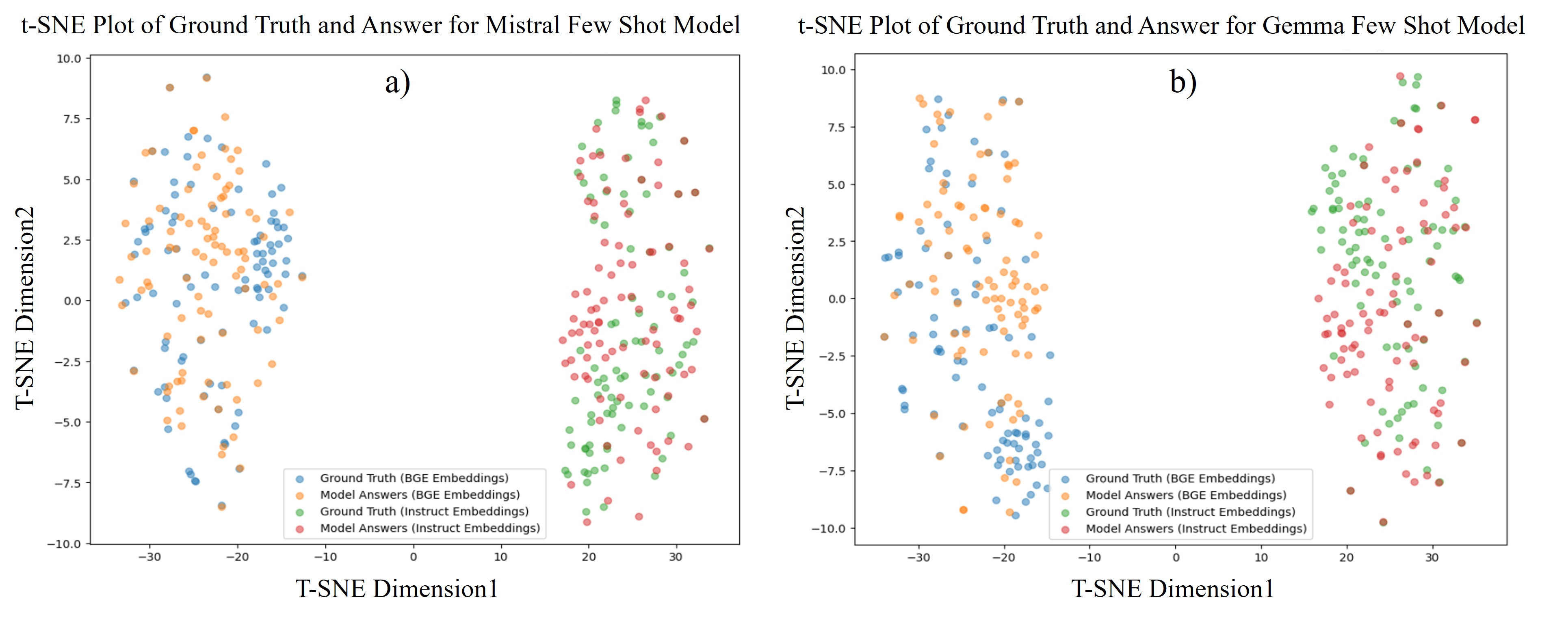}
    \caption{Figures 8.a and 8.b: t-SNE plot of ground truth and model answers for Mistral and Gemma using few shots prompt.}
    \label{fig8}
\end{figure}

In the Mistral Few-Shot Model, the clustering of ground truth and model answers appears to be more compact, particularly for Instruct Embeddings. This suggests that the model may have an improved ability to match medical terminology when provided with additional examples. However, there are still areas of the plot where embeddings do not overlap entirely, indicating some semantic disparities. Nonetheless, the cosine similarity between the two vectors, utilizing both embeddings, demonstrates significant results, affirming the model's capability to produce answers closely aligned with the ground truth. Similarly, the t-SNE visualization for the Gemma Few-Shot Model also indicates a trend towards tighter clustering between the ground truth and model answers. This underscores the notion that with an increased number of examples, the model becomes more adept at capturing the nuanced language of the ground truth.

\begin{figure}[h]
\centering
        \includegraphics[totalheight=6cm]{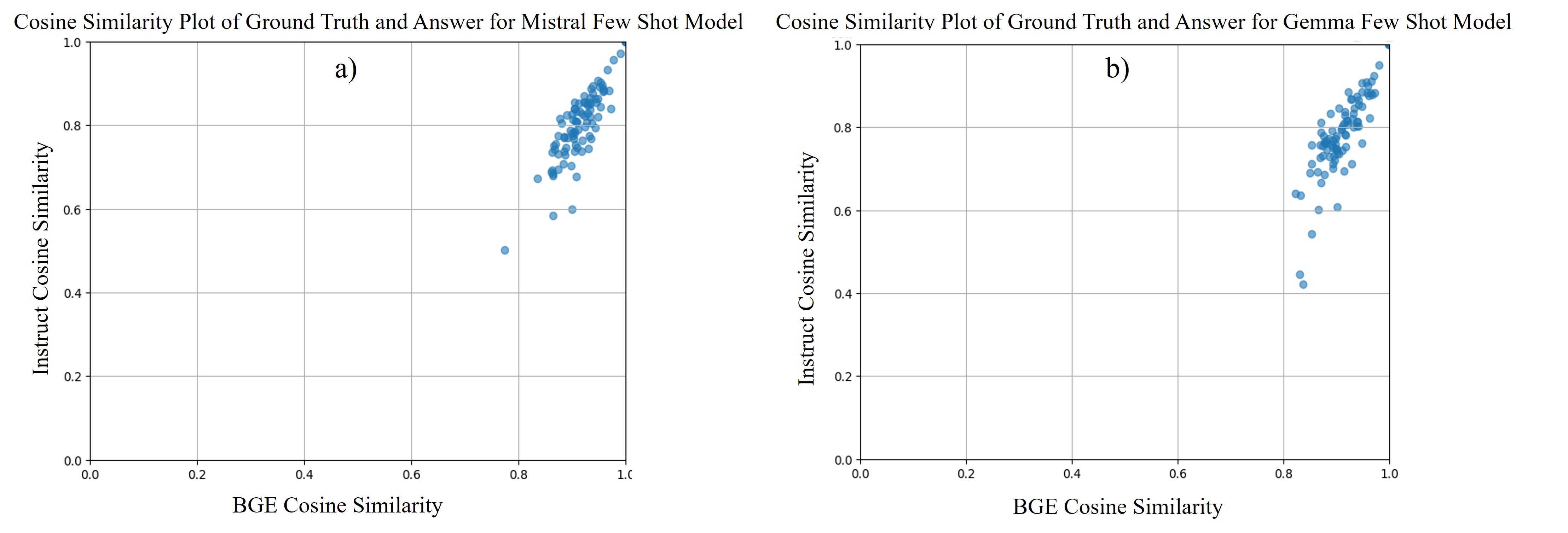}
    \caption{Figures 9.a and 9.b: Cosine similarity plot for Gemma and Mistral models using few shots prompt.}
    \label{fig9}
\end{figure}

\section{Conclusions and Future Work}

To explore leveraging open-source LLMs for medical textual data information extraction, we have demonstrated the effectiveness of GAMedX, a prompt engineering-based unified approach where pydantic schema is employed. GAMedX is a LLM agnostic approach and for experimentation purposes, Mistral 7B and Gemma 7B were resorted to. Additionally, to prove the versatility of the approach when it comes to extracting information from unstructured textual healthcare data, two datasets were used for retrieving various patient-related information in a structured format and reliable manner. Thus, enabling the potential use of GAMedX in various healthcare applications. The experiments are followed by a comprehensive analysis of the approach’s performance showing its robustness across different contexts. The evaluation comprises two parts, the first one consists of a quantitative analysis through ROUGE scores, carried out for both datasets, while the second one is a semantics analysis conducted for VAERS dataset and utilizing t-SNE with both BGE and Instruct Embeddings. Both analyses, highlight the significant potential of LLMs in enhancing useful and accurate information extraction from medical textual data. 
\\[0.25cm]
Moving forward, we aim to refine our proposed approach by exploring other open-source LLMs, with a particular focus on expanding our methodology to identify which LLM integration corresponds more conveniently to which type of healthcare textual data. Furthermore, investigating various NLP tasks relevant in healthcare (e.g., sentiment analysis) present an exciting venue for future research. GAMedX applications and potential enhancements not only contributes academically to the research revolving around open-source LLMs application, but also constitutes a practical tool able to automate multiple auxiliary tasks in healthcare, and therefore save healthcare professional’s time for those who need it the most: the patients.



\begin{thebibliography}{10}

\bibitem{bush2017structured}
Ruth~A Bush, Cynthia Kuelbs, Julie Ryu, Wen Jiang, and George Chiang.
\newblock Structured data entry in the electronic medical record: perspectives of pediatric specialty physicians and surgeons.
\newblock {\em Journal of medical systems}, 41:1--8, 2017.

\bibitem{meystre2008extracting}
St{\'e}phane~M Meystre, Guergana~K Savova, Karin~C Kipper-Schuler, and John~F Hurdle.
\newblock Extracting information from textual documents in the electronic health record: a review of recent research.
\newblock {\em Yearbook of medical informatics}, 17(01):128--144, 2008.

\bibitem{liang2019evaluation}
Huiying Liang, Brian~Y Tsui, Hao Ni, Carolina~CS Valentim, Sally~L Baxter, Guangjian Liu, Wenjia Cai, Daniel~S Kermany, Xin Sun, Jiancong Chen, et~al.
\newblock Evaluation and accurate diagnoses of pediatric diseases using artificial intelligence.
\newblock {\em Nature medicine}, 25(3):433--438, 2019.

\bibitem{yang2021assessing}
Jie Yang, John~W Lian, Yen-Po~Harvey Chin, Liqin Wang, Anna Lian, George~F Murphy, and Li~Zhou.
\newblock Assessing the prognostic significance of tumor-infiltrating lymphocytes in patients with melanoma using pathologic features identified by natural language processing.
\newblock {\em JAMA Network Open}, 4(9):e2126337--e2126337, 2021.

\bibitem{nadkarni2011natural}
Prakash~M Nadkarni, Lucila Ohno-Machado, and Wendy~W Chapman.
\newblock Natural language processing: an introduction.
\newblock {\em Journal of the American Medical Informatics Association}, 18(5):544--551, 2011.

\bibitem{zheng2011coreference}
Jiaping Zheng, Wendy~W Chapman, Rebecca~S Crowley, and Guergana~K Savova.
\newblock Coreference resolution: A review of general methodologies and applications in the clinical domain.
\newblock {\em Journal of biomedical informatics}, 44(6):1113--1122, 2011.

\bibitem{drolet2017electronic}
Brian~C Drolet, Jayson~S Marwaha, Brad Hyatt, Phillip~E Blazar, and Scott~D Lifchez.
\newblock Electronic communication of protected health information: privacy, security, and hipaa compliance.
\newblock {\em The Journal of hand surgery}, 42(6):411--416, 2017.

\bibitem{reddy2023evaluating}
Sandeep Reddy.
\newblock Evaluating large language models for use in healthcare: A framework for translational value assessment.
\newblock {\em Informatics in Medicine Unlocked}, page 101304, 2023.

\bibitem{yang2023large}
Rui Yang, Ting~Fang Tan, Wei Lu, Arun~James Thirunavukarasu, Daniel Shu~Wei Ting, and Nan Liu.
\newblock Large language models in health care: Development, applications, and challenges.
\newblock {\em Health Care Science}, 2(4):255--263, 2023.

\bibitem{xia2012clinical}
Fei Xia and Meliha Yetisgen-Yildiz.
\newblock Clinical corpus annotation: challenges and strategies.
\newblock In {\em Proceedings of the third workshop on building and evaluating resources for biomedical text mining (BioTxtM’2012) in conjunction with the international conference on language resources and evaluation (LREC), Istanbul, Turkey}, pages 21--27, 2012.

\bibitem{zhang2020bert}
Zachariah Zhang, Jingshu Liu, and Narges Razavian.
\newblock Bert-xml: Large scale automated icd coding using bert pretraining.
\newblock {\em arXiv preprint arXiv:2006.03685}, 2020.

\bibitem{si2019deep}
Yuqi Si and Kirk Roberts.
\newblock Deep patient representation of clinical notes via multi-task learning for mortality prediction.
\newblock {\em AMIA Summits on Translational Science Proceedings}, 2019:779, 2019.

\bibitem{chakraborty2024need}
Chiranjib Chakraborty, Manojit Bhattacharya, and Sang-Soo Lee.
\newblock Need an ai-enabled, next-generation, advanced chatgpt or large language models (llms) for error-free and accurate medical information.
\newblock {\em Annals of Biomedical Engineering}, 52(2):134--135, 2024.

\bibitem{lecun2015deep}
Yann LeCun, Yoshua Bengio, and Geoffrey Hinton.
\newblock Deep learning.
\newblock {\em nature}, 521(7553):436--444, 2015.

\bibitem{collobert2011natural}
Ronan Collobert, Jason Weston, L{\'e}on Bottou, Michael Karlen, Koray Kavukcuoglu, and Pavel Kuksa.
\newblock Natural language processing (almost) from scratch.
\newblock {\em Journal of machine learning research}, 12:2493--2537, 2011.

\bibitem{lample2016neural}
Guillaume Lample, Miguel Ballesteros, Sandeep Subramanian, Kazuya Kawakami, and Chris Dyer.
\newblock Neural architectures for named entity recognition.
\newblock {\em arXiv preprint arXiv:1603.01360}, 2016.

\bibitem{vaswani2017attention}
Ashish Vaswani, Noam Shazeer, Niki Parmar, Jakob Uszkoreit, Llion Jones, Aidan~N Gomez, {\L}ukasz Kaiser, and Illia Polosukhin.
\newblock Attention is all you need.
\newblock {\em Advances in neural information processing systems}, 30, 2017.

\bibitem{devlin2018bert}
Jacob Devlin, Ming-Wei Chang, Kenton Lee, and Kristina Toutanova.
\newblock Bert: Pre-training of deep bidirectional transformers for language understanding.
\newblock {\em arXiv preprint arXiv:1810.04805}, 2018.

\bibitem{yu2020named}
Juntao Yu, Bernd Bohnet, and Massimo Poesio.
\newblock Named entity recognition as dependency parsing.
\newblock {\em arXiv preprint arXiv:2005.07150}, 2020.

\bibitem{yamada2020luke}
Ikuya Yamada, Akari Asai, Hiroyuki Shindo, Hideaki Takeda, and Yuji Matsumoto.
\newblock Luke: Deep contextualized entity representations with entity-aware self-attention.
\newblock {\em arXiv preprint arXiv:2010.01057}, 2020.

\bibitem{lyu2021relation}
Shengfei Lyu and Huanhuan Chen.
\newblock Relation classification with entity type restriction.
\newblock {\em arXiv preprint arXiv:2105.08393}, 2021.

\bibitem{ye2021packed}
Deming Ye, Yankai Lin, Peng Li, and Maosong Sun.
\newblock Packed levitated marker for entity and relation extraction.
\newblock {\em arXiv preprint arXiv:2109.06067}, 2021.

\bibitem{xu2021entity}
Benfeng Xu, Quan Wang, Yajuan Lyu, Yong Zhu, and Zhendong Mao.
\newblock Entity structure within and throughout: Modeling mention dependencies for document-level relation extraction.
\newblock In {\em Proceedings of the AAAI conference on artificial intelligence}, volume~35, pages 14149--14157, 2021.

\bibitem{raffel2020exploring}
Colin Raffel, Noam Shazeer, Adam Roberts, Katherine Lee, Sharan Narang, Michael Matena, Yanqi Zhou, Wei Li, and Peter~J Liu.
\newblock Exploring the limits of transfer learning with a unified text-to-text transformer.
\newblock {\em Journal of machine learning research}, 21(140):1--67, 2020.

\bibitem{jiang2019smart}
Haoming Jiang, Pengcheng He, Weizhu Chen, Xiaodong Liu, Jianfeng Gao, and Tuo Zhao.
\newblock Smart: Robust and efficient fine-tuning for pre-trained natural language models through principled regularized optimization.
\newblock {\em arXiv preprint arXiv:1911.03437}, 2019.

\bibitem{yang2019xlnet}
Zhilin Yang, Zihang Dai, Yiming Yang, Jaime Carbonell, Russ~R Salakhutdinov, and Quoc~V Le.
\newblock Xlnet: Generalized autoregressive pretraining for language understanding.
\newblock {\em Advances in neural information processing systems}, 32, 2019.

\bibitem{zhang2020semantics}
Zhuosheng Zhang, Yuwei Wu, Hai Zhao, Zuchao Li, Shuailiang Zhang, Xi~Zhou, and Xiang Zhou.
\newblock Semantics-aware bert for language understanding.
\newblock In {\em Proceedings of the AAAI Conference on Artificial Intelligence}, volume~34, pages 9628--9635, 2020.

\bibitem{lan2019albert}
Zhenzhong Lan, Mingda Chen, Sebastian Goodman, Kevin Gimpel, Piyush Sharma, and Radu Soricut.
\newblock Albert: A lite bert for self-supervised learning of language representations.
\newblock {\em arXiv preprint arXiv:1909.11942}, 2019.

\bibitem{zhang2021retrospective}
Zhuosheng Zhang, Junjie Yang, and Hai Zhao.
\newblock Retrospective reader for machine reading comprehension.
\newblock In {\em Proceedings of the AAAI conference on artificial intelligence}, volume~35, pages 14506--14514, 2021.

\bibitem{garg2020tanda}
Siddhant Garg, Thuy Vu, and Alessandro Moschitti.
\newblock Tanda: Transfer and adapt pre-trained transformer models for answer sentence selection.
\newblock In {\em Proceedings of the AAAI conference on artificial intelligence}, volume~34, pages 7780--7788, 2020.

\bibitem{thirunavukarasu2023large}
Arun~James Thirunavukarasu, Darren Shu~Jeng Ting, Kabilan Elangovan, Laura Gutierrez, Ting~Fang Tan, and Daniel Shu~Wei Ting.
\newblock Large language models in medicine.
\newblock {\em Nature medicine}, 29(8):1930--1940, 2023.

\bibitem{zhang2020using}
Mengyuan Zhang, Jin Wang, and Xuejie Zhang.
\newblock Using a pre-trained language model for medical named entity extraction in chinese clinic text.
\newblock In {\em 2020 IEEE 10th International Conference on Electronics Information and Emergency Communication (ICEIEC)}, pages 312--317. IEEE, 2020.

\bibitem{wu2020deep}
Stephen Wu, Kirk Roberts, Surabhi Datta, Jingcheng Du, Zongcheng Ji, Yuqi Si, Sarvesh Soni, Qiong Wang, Qiang Wei, Yang Xiang, et~al.
\newblock Deep learning in clinical natural language processing: a methodical review.
\newblock {\em Journal of the American Medical Informatics Association}, 27(3):457--470, 2020.

\bibitem{bommasani2021opportunities}
Rishi Bommasani, Drew~A Hudson, Ehsan Adeli, Russ Altman, Simran Arora, Sydney von Arx, Michael~S Bernstein, Jeannette Bohg, Antoine Bosselut, Emma Brunskill, et~al.
\newblock On the opportunities and risks of foundation models.
\newblock {\em arXiv preprint arXiv:2108.07258}, 2021.

\bibitem{floridi2020gpt}
Luciano Floridi and Massimo Chiriatti.
\newblock Gpt-3: Its nature, scope, limits, and consequences.
\newblock {\em Minds and Machines}, 30:681--694, 2020.

\bibitem{Gu2021Biomedical}
Yu~Gu, Tinn Robert, Cheng Hao, Lucas Michael, Usuyama Naoto, Liu Xiaodong, Naumann Tristan, Gao Jianfeng, and Poon Hoifung.
\newblock Domain-specific language model pretraining for biomedical natural language processing.
\newblock {\em ACM Transactions on Computing for Healthcare (HEALTH)}, 3(1):1--23, 2021.

\bibitem{shin-etal-2020-biomegatron}
Hoo-Chang Shin, Yang Zhang, Evelina Bakhturina, Raul Puri, Mostofa Patwary, Mohammad Shoeybi, and Raghav Mani.
\newblock {B}io{M}egatron: Larger biomedical domain language model.
\newblock In Bonnie Webber, Trevor Cohn, Yulan He, and Yang Liu, editors, {\em Proceedings of the 2020 Conference on Empirical Methods in Natural Language Processing (EMNLP)}, pages 4700--4706, Online, November 2020. Association for Computational Linguistics.

\bibitem{alsentzer-etal-2019-publicly}
Emily Alsentzer, John Murphy, William Boag, Wei-Hung Weng, Di~Jindi, Tristan Naumann, and Matthew McDermott.
\newblock Publicly available clinical {BERT} embeddings.
\newblock In Anna Rumshisky, Kirk Roberts, Steven Bethard, and Tristan Naumann, editors, {\em Proceedings of the 2nd Clinical Natural Language Processing Workshop}, pages 72--78, Minneapolis, Minnesota, USA, June 2019. Association for Computational Linguistics.

\bibitem{yang2022large}
Xi~Yang, Aokun Chen, Nima PourNejatian, Hoo~Chang Shin, Kaleb~E Smith, Christopher Parisien, Colin Compas, Cheryl Martin, Anthony~B Costa, Mona~G Flores, et~al.
\newblock A large language model for electronic health records.
\newblock {\em NPJ digital medicine}, 5(1):194, 2022.

\bibitem{arora2023promise}
Anmol Arora and Ananya Arora.
\newblock The promise of large language models in health care.
\newblock {\em The Lancet}, 401(10377):641, 2023.

\bibitem{mann2020language}
Ben Mann, N~Ryder, M~Subbiah, J~Kaplan, P~Dhariwal, A~Neelakantan, P~Shyam, G~Sastry, A~Askell, S~Agarwal, et~al.
\newblock Language models are few-shot learners.
\newblock {\em arXiv preprint arXiv:2005.14165}, 2020.

\bibitem{liu2023pre}
Pengfei Liu, Weizhe Yuan, Jinlan Fu, Zhengbao Jiang, Hiroaki Hayashi, and Graham Neubig.
\newblock Pre-train, prompt, and predict: A systematic survey of prompting methods in natural language processing.
\newblock {\em ACM Computing Surveys}, 55(9):1--35, 2023.

\bibitem{sanh2021multitask}
Victor Sanh, Albert Webson, Colin Raffel, Stephen~H Bach, Lintang Sutawika, Zaid Alyafeai, Antoine Chaffin, Arnaud Stiegler, Teven~Le Scao, Arun Raja, et~al.
\newblock Multitask prompted training enables zero-shot task generalization.
\newblock {\em arXiv preprint arXiv:2110.08207}, 2021.

\bibitem{mishra2021reframing}
Swaroop Mishra, Daniel Khashabi, Chitta Baral, Yejin Choi, and Hannaneh Hajishirzi.
\newblock Reframing instructional prompts to gptk's language.
\newblock {\em arXiv preprint arXiv:2109.07830}, 2021.

\bibitem{yang2022gpt}
Xiaohan Yang, Eduardo Peynetti, Vasco Meerman, and Chris Tanner.
\newblock What gpt knows about who is who.
\newblock {\em arXiv preprint arXiv:2205.07407}, 2022.

\bibitem{liu2022qaner}
Andy~T Liu, Wei Xiao, Henghui Zhu, Dejiao Zhang, Shang-Wen Li, and Andrew Arnold.
\newblock Qaner: Prompting question answering models for few-shot named entity recognition.
\newblock {\em arXiv preprint arXiv:2203.01543}, 2022.

\bibitem{yang2020identifying}
Xi~Yang, Jiang Bian, Ruogu Fang, Ragnhildur~I Bjarnadottir, William~R Hogan, and Yonghui Wu.
\newblock Identifying relations of medications with adverse drug events using recurrent convolutional neural networks and gradient boosting.
\newblock {\em Journal of the American Medical Informatics Association}, 27(1):65--72, 2020.

\bibitem{yang2019madex}
Xi~Yang, Jiang Bian, Yan Gong, William~R Hogan, and Yonghui Wu.
\newblock Madex: a system for detecting medications, adverse drug events, and their relations from clinical notes.
\newblock {\em Drug safety}, 42:123--133, 2019.

\bibitem{yang2019study}
Xi~Yang, Tianchen Lyu, Qian Li, Chih-Yin Lee, Jiang Bian, William~R Hogan, and Yonghui Wu.
\newblock A study of deep learning methods for de-identification of clinical notes in cross-institute settings.
\newblock {\em BMC medical informatics and decision making}, 19:1--9, 2019.

\bibitem{ferraro2017effects}
Jeffrey~P Ferraro, Ye~Ye, Per~H Gesteland, Peter~J Haug, Fuchiang Tsui, Gregory~F Cooper, Rudy Van~Bree, Thomas Ginter, Andrew~J Nowalk, and Michael Wagner.
\newblock The effects of natural language processing on cross-institutional portability of influenza case detection for disease surveillance.
\newblock {\em Applied clinical informatics}, 8(02):560--580, 2017.

\bibitem{peng2023clinical}
Cheng Peng, Xi~Yang, Zehao Yu, Jiang Bian, William~R Hogan, and Yonghui Wu.
\newblock Clinical concept and relation extraction using prompt-based machine reading comprehension.
\newblock {\em Journal of the American Medical Informatics Association}, 30(9):1486--1493, 2023.

\bibitem{chen2023contextualized}
Aokun Chen, Zehao Yu, Xi~Yang, Yi~Guo, Jiang Bian, and Yonghui Wu.
\newblock Contextualized medication information extraction using transformer-based deep learning architectures.
\newblock {\em Journal of biomedical informatics}, 142:104370, 2023.

\bibitem{henry20202018}
Sam Henry, Kevin Buchan, Michele Filannino, Amber Stubbs, and Ozlem Uzuner.
\newblock 2018 n2c2 shared task on adverse drug events and medication extraction in electronic health records.
\newblock {\em Journal of the American Medical Informatics Association}, 27(1):3--12, 2020.

\bibitem{lybarger20232022}
Kevin Lybarger, Meliha Yetisgen, and {\"O}zlem Uzuner.
\newblock The 2022 n2c2/uw shared task on extracting social determinants of health.
\newblock {\em Journal of the American Medical Informatics Association}, 30(8):1367--1378, 2023.

\bibitem{li2019unified}
Xiaoya Li, Jingrong Feng, Yuxian Meng, Qinghong Han, Fei Wu, and Jiwei Li.
\newblock A unified mrc framework for named entity recognition.
\newblock {\em arXiv preprint arXiv:1910.11476}, 2019.

\bibitem{ouyang2022training}
Long Ouyang, Jeffrey Wu, Xu~Jiang, Diogo Almeida, Carroll Wainwright, Pamela Mishkin, Chong Zhang, Sandhini Agarwal, Katarina Slama, Alex Ray, et~al.
\newblock Training language models to follow instructions with human feedback.
\newblock {\em Advances in neural information processing systems}, 35:27730--27744, 2022.

\bibitem{agrawal2022large}
Monica Agrawal, Stefan Hegselmann, Hunter Lang, Yoon Kim, and David Sontag.
\newblock Large language models are few-shot clinical information extractors.
\newblock {\em arXiv preprint arXiv:2205.12689}, 2022.

\bibitem{ahmed2024med}
Awais Ahmed, Xiaoyang Zeng, Rui Xi, Mengshu Hou, and Syed~Attique Shah.
\newblock Med-prompt: A novel prompt engineering framework for medicine prediction on free-text clinical notes.
\newblock {\em Journal of King Saud University-Computer and Information Sciences}, 36(2):101933, 2024.

\bibitem{cdc_vaers}
{US Centers for Disease Control and Prevention}.
\newblock {VAERS - About Us}.
\newblock \url{https://vaers.hhs.gov/about.html}.
\newblock Accessed May 15, 2024.

\bibitem{li2024ae}
Yiming Li, Jianfu Li, Jianping He, and Cui Tao.
\newblock Ae-gpt: Using large language models to extract adverse events from surveillance reports-a use case with influenza vaccine adverse events.
\newblock {\em Plos one}, 19(3):e0300919, 2024.

\bibitem{jiang2023mistral}
Albert~Q Jiang, Alexandre Sablayrolles, Arthur Mensch, Chris Bamford, Devendra~Singh Chaplot, Diego de~las Casas, Florian Bressand, Gianna Lengyel, Guillaume Lample, Lucile Saulnier, et~al.
\newblock Mistral 7b.
\newblock {\em arXiv preprint arXiv:2310.06825}, 2023.

\bibitem{team2024gemma}
Gemma Team, Thomas Mesnard, Cassidy Hardin, Robert Dadashi, Surya Bhupatiraju, Shreya Pathak, Laurent Sifre, Morgane Rivi{\`e}re, Mihir~Sanjay Kale, Juliette Love, et~al.
\newblock Gemma: Open models based on gemini research and technology.
\newblock {\em arXiv preprint arXiv:2403.08295}, 2024.

\bibitem{van2008visualizing}
Laurens Van~der Maaten and Geoffrey Hinton.
\newblock Visualizing data using t-sne.
\newblock {\em Journal of machine learning research}, 9(11), 2008.

\end{thebibliography}
\end{document}